%% file: main.tex
\documentclass[10pt,twocolumn,letterpaper]{article}

\usepackage[pagenumbers]{cvpr} % To force page numbers, e.g. for an arXiv version
\usepackage{multirow}
\input{preamble}
\input{tabs}

\input{figs}

\definecolor{cvprblue}{rgb}{0.21,0.49,0.74}
\usepackage[pagebackref,breaklinks,colorlinks,allcolors=cvprblue]{hyperref}

\def\paperID{2} % *** Enter the Paper ID here
\def\confName{CVPR}
\def\confYear{2026}

\title{HORNet: Task-Guided Frame Selection for Video Question Answering with Vision-Language Models}
\author{Xiangyu Bai$^*$ \quad Bishoy Galoaa$^*$ \quad Sarah Ostadabbas\\
Northeastern University\\
Boston, MA, USA\\
{\tt\small \{bai.xiang, galoaa.b, s.ostadabbas\}@northeastern.edu}\\
$^*$These authors contributed equally to this work.
}

\begin{document}
\maketitle

\input{sec/0_abstract}    
\input{sec/1_intro}
\input{sec/2_related}

\input{sec/3_method}
\input{sec/4_results}
\input{sec/5_conclusion}
{
    \small
    \bibliographystyle{ieeenat_fullname}
    \bibliography{main}
}


\end{document}

%% file: preamble.tex
\usepackage{booktabs}
\usepackage{xcolor}
\usepackage{colortbl}
\usepackage{makecell}
\definecolor{hornetgold}{RGB}{255, 195, 0}
\definecolor{rowgray}{RGB}{245,245,245}
\definecolor{darkgreen}{RGB}{34,139,34}
\definecolor{darkred}{RGB}{180,30,30}

\DeclareMathOperator*{\argmax}{arg\,max}

%% file: tabs.tex
\definecolor{hornetgold}{RGB}{255, 195, 0}
\definecolor{rowgray}{RGB}{245,245,245}
\definecolor{mygreen}{RGB}{34,139,34}
\definecolor{myred}{RGB}{180,30,30}

\newcommand{\cyes}{\textcolor{mygreen}{\textbf{\checkmark}}}
\newcommand{\cno}{\textcolor{myred}{$\boldsymbol{\times}$}}
\newcommand{\cprt}{\textcolor{orange}{${\sim}$}}

\newcommand{\researchgap}{%
\begin{table}[t]
\centering
\caption{\textbf{Research gap.} Existing frame selection methods satisfy at most two of four desirable properties simultaneously. HORNet is the first to achieve all four: learned selection, reward-based optimization, a fully frozen VLM, and parameter efficiency ($<$1M parameters). \cyes~fully supported, \cno~not supported, \cprt~partial.
}
\label{tab:gap}
\footnotesize
\begin{tabular}{lcccc}
\toprule
\textbf{Method} &
\makecell{\textbf{Learned}\\\textbf{Selection}} &
\makecell{\textbf{Reward}\\\textbf{Optimized}} &
\makecell{\textbf{Frozen}\\\textbf{VLM}} &
\makecell{\textbf{Param.}\\\textbf{Efficient}} \\
\midrule
Uniform Sampling                            & \cno  & \cno  & \cyes & \cyes \\
\rowcolor{rowgray}
SeViLA~\cite{sevila}                        & \cyes & \cno  & \cno  & \cno  \\
Frame-Voyager~\cite{framevoyager}           & \cyes & \cno  & \cno  & \cno  \\
\rowcolor{rowgray}
F2C~\cite{f2c2025}                          & \cno  & \cno  & \cyes & \cyes \\
ReFoCUS~\cite{refocus2025}                  & \cyes & \cyes & \cprt & \cno  \\
\rowcolor{rowgray}
ViaRL~\cite{viarl2025}                      & \cyes & \cyes & \cno  & \cno  \\
\midrule
\rowcolor{hornetgold}
\textbf{HORNet (Ours)}                      & \cyes & \cyes & \cyes & \cyes \\
\bottomrule
\end{tabular}
\end{table}
}

\newcommand{\mainoe}{%
\begin{table*}[t]
\centering
\caption{Performance comparison across open-ended QA datasets. We adopt the aforementioned F1-Lev metric to measure model performance, which is a  weighted combination of token-level F1 and normalized edit similarity on lemmatized texts. We also report efficiency measured in runtime and average frames passed to Qwen. Qwen’s baseline processing time includes uniform sampling, video encoding, and answer generation. In our setup, frame selection replaces the sampling step and is reported separately. Even under this accounting, the combined runtime of HORNet still yields a notable overall speedup. Additionally, when comparing generation speed, we follow Qwen’s default sampling rate (fps=2). Under the assumption of a 24-fps source video, the baseline effectively processes roughly 1/12 of all frames-still substantially more input than HORNet requires. We highlight best performance for each benchmark in \textbf{bold} and mark performance \textcolor{mygreen}{gain} or \textcolor{myred}{loss} as percentages.}
\label{tab:mainoe}
\footnotesize
\begin{tabular}{l l c c c c}
\toprule
\textbf{Dataset} & \textbf{Model} & \textbf{F1-Lev$\uparrow$} & \textbf{Frame Sel. (s)$\downarrow$} & \textbf{Qwen Proc. (s)$\downarrow$} & \textbf{Avg. Frames$\downarrow$} \\
\midrule
\multirow{2}{*}{MSVD~\cite{msvdqa}}
    & Qwen3-VL-2B (Baseline)     & 0.3483 & --   & 0.28          & 11.65 \\
    & \cellcolor{rowgray}HORNet+Qwen3-VL-2B (Ours)  & \cellcolor{rowgray}\textbf{0.3543} \textcolor{mygreen}{\scriptsize +1.7\%} & \cellcolor{rowgray}0.12 & \cellcolor{rowgray}\textbf{0.10} \textcolor{mygreen}{\scriptsize $\downarrow$64\%} & \cellcolor{rowgray}\textbf{4.00} \textcolor{mygreen}{\scriptsize $\downarrow$66\%} \\
\midrule
\multirow{2}{*}{MSRVTT~\cite{msrvttqa}}
    & Qwen3-VL-2B (Baseline)     & \textbf{0.3209} & --   & 0.58          & 47.52 \\
    & \cellcolor{rowgray}HORNet+Qwen3-VL-2B (Ours)  & \cellcolor{rowgray}0.3029 \textcolor{myred}{\scriptsize -5.6\%} & \cellcolor{rowgray}0.09 & \cellcolor{rowgray}\textbf{0.09} \textcolor{mygreen}{\scriptsize $\downarrow$84\%} & \cellcolor{rowgray}\textbf{4.00} \textcolor{mygreen}{\scriptsize $\downarrow$92\%} \\
\midrule
\multirow{2}{*}{NextOE~\cite{nextqa}}
    & Qwen3-VL-2B (Baseline)     & \textbf{0.3045} & --   & 1.01          & 1157.88 \\
    & \cellcolor{rowgray}HORNet+Qwen3-VL-2B (Ours)  & \cellcolor{rowgray}0.2738 \textcolor{myred}{\scriptsize -10.1\%} & \cellcolor{rowgray}0.52 & \cellcolor{rowgray}\textbf{0.19} \textcolor{mygreen}{\scriptsize $\downarrow$81\%} & \cellcolor{rowgray}\textbf{8.00} \textcolor{mygreen}{\scriptsize $\downarrow$99\%} \\
\bottomrule
\end{tabular}
\end{table*}
}

\newcommand{\mainmcq}{%
\begin{table*}[t]
\centering
\caption{Performance comparison across selection-based MCQ datasets. We adopt selection-accuracy as our metric and report efficiency gain of HORNet similar to Table \ref{tab:mainoe}. For each of these dataset we randomly sampled 1,000 QA pairs.}
\label{tab:mainmcq}
\footnotesize
\begin{tabular}{l l c c c c}
\toprule
\textbf{Dataset} & \textbf{Model} & \textbf{Accuracy(\%)$\uparrow$} & \textbf{Frame Sel. (s)$\downarrow$} & \textbf{Qwen Proc. (s)$\downarrow$} & \textbf{Avg. Frames$\downarrow$} \\
\midrule
\multirow{2}{*}{VideoMME~\cite{videomme}}
    & Qwen3-VL-2B (Baseline)     & \textbf{68.30} & --   & 2.53          & 3066.73 \\
    & \cellcolor{rowgray}HORNet+Qwen3-VL-2B (Ours)  & \cellcolor{rowgray}52.10 \textcolor{myred}{\scriptsize -16.2\%} & \cellcolor{rowgray}1.51 & \cellcolor{rowgray}\textbf{0.18} \textcolor{mygreen}{\scriptsize $\downarrow$93\%} & \cellcolor{rowgray}\textbf{8.00} \textcolor{mygreen}{\scriptsize $\downarrow$99\%} \\
\midrule
\multirow{2}{*}{ActivityNetQA~\cite{activityqa}}
    & Qwen3-VL-2B (Baseline)     & \textbf{75.00} & --   & 2.37          & 3152.49 \\
    & \cellcolor{rowgray}HORNet+Qwen3-VL-2B (Ours)  & \cellcolor{rowgray}68.80 \textcolor{myred}{\scriptsize -6.2\%} & \cellcolor{rowgray}1.64 & \cellcolor{rowgray}\textbf{0.17} \textcolor{mygreen}{\scriptsize $\downarrow$93\%} & \cellcolor{rowgray}\textbf{8.00} \textcolor{mygreen}{\scriptsize $\downarrow$99\%} \\
\midrule
\multirow{2}{*}{NextQA~\cite{nextqa}}
    & Qwen3-VL-2B (Baseline)     & \textbf{76.80} & --   & 0.98          & 1157.88 \\
    & \cellcolor{rowgray}HORNet+Qwen3-VL-2B (Ours)  & \cellcolor{rowgray}71.50 \textcolor{myred}{\scriptsize -5.3\%} & \cellcolor{rowgray}0.53 & \cellcolor{rowgray}\textbf{0.25} \textcolor{mygreen}{\scriptsize $\downarrow$74\%} & \cellcolor{rowgray}\textbf{8.00} \textcolor{mygreen}{\scriptsize $\downarrow$99\%} \\
\bottomrule
\end{tabular}
\end{table*}
}

\newcommand{\ablationobjective}{%
\begin{table}[t]
\centering
\caption{
    \textbf{Training objective ablation.}
    All variants use the same TimeSformer-Tiny encoder, MLP policy ($<$1M params),
    and frozen Qwen3-VL-2B answerer. Trained on MSVD-QA only.
    MSRVTT-QA results show out-of-distribution generalization.
}
\label{tab:ablation_objective}
\setlength{\tabcolsep}{5pt}
\small
\begin{tabular}{lcc}
\toprule
\textbf{Training} & \makecell{\textbf{MSVD}\\\textbf{(F1-Lev$\uparrow$)}} & \makecell{\textbf{MSRVTT}\\\textbf{(F1-Lev$\uparrow$)}} \\
\midrule
No training (baseline) & 0.3483 & \textbf{0.3209} \\
SFT (weighted BCE)     & 0.3495 & 0.2882 \\
PPO (clipped surrogate)& \textbf{0.3585} & 0.2948 \\
\cellcolor{rowgray}\textbf{GRPO (Ours)}   & \cellcolor{rowgray}0.3543 & \cellcolor{rowgray}0.3029 \\
\bottomrule
\end{tabular}
\end{table}
}

\newcommand{\ablationselection}{%
\begin{table}[t]
\centering
\caption{
    \textbf{Frame selection strategy ablation.}
    All methods select 4 frames and pass them to a frozen Qwen3-VL-2B answerer.
    MSVD-QA and MSRVTT-QA report F1-Lev; NExT-QA reports MCQ accuracy (\%).
}
\label{tab:ablation_selection}
\setlength{\tabcolsep}{4pt}
\small
\begin{tabular}{lccc}
\toprule
\textbf{Strategy} & \makecell{\textbf{MSVD}\\\textbf{(F1-Lev$\uparrow$)}} & \makecell{\textbf{MSRVTT}\\\textbf{(F1-Lev$\uparrow$)}} & \makecell{\textbf{NExT-QA}\\\textbf{(Acc.$\uparrow$)}} \\
\midrule
Random             & 0.3527 & 0.3027 & 65.88 \\

Uniform            & 0.3493 & \textbf{0.3058} & 64.24 \\
\cellcolor{rowgray}\textbf{HORNet}    & \cellcolor{rowgray}\textbf{0.3543} & \cellcolor{rowgray}0.3029 & \cellcolor{rowgray}\textbf{71.50} \\
\bottomrule
\end{tabular}
\end{table}
}

\newcommand{\ablationvlm}{%
\begin{table}[t]
\centering
\caption{
    \textbf{VLM answerer ablation on MSVD-QA.}
    The same HORNet policy (trained with GRPO) selects frames.
    Only the frozen answerer is swapped; frame selection is identical.
}
\label{tab:ablation_vlm}
\setlength{\tabcolsep}{5pt}
\small
\begin{tabular}{llc}
\toprule
\textbf{VLM Answerer} & \textbf{Size} & \textbf{F1-Lev$\uparrow$} \\
\midrule
Qwen3-VL-Instruct (baseline)       & 2B & 0.3483 \\
Qwen3-VL-Instruct + HORNet         & 2B & 0.3543 \\
\midrule
\cellcolor{rowgray} Qwen2.5-VL-Instruct + HORNet       & \cellcolor{rowgray}3B & \cellcolor{rowgray}\textbf{0.3846} \\
\bottomrule
\end{tabular}
\end{table}
}

%% file: figs.tex
\newcommand{\pipelinefig}{%
\begin{figure*}[t]
    \centering
    \includegraphics[width=\linewidth]{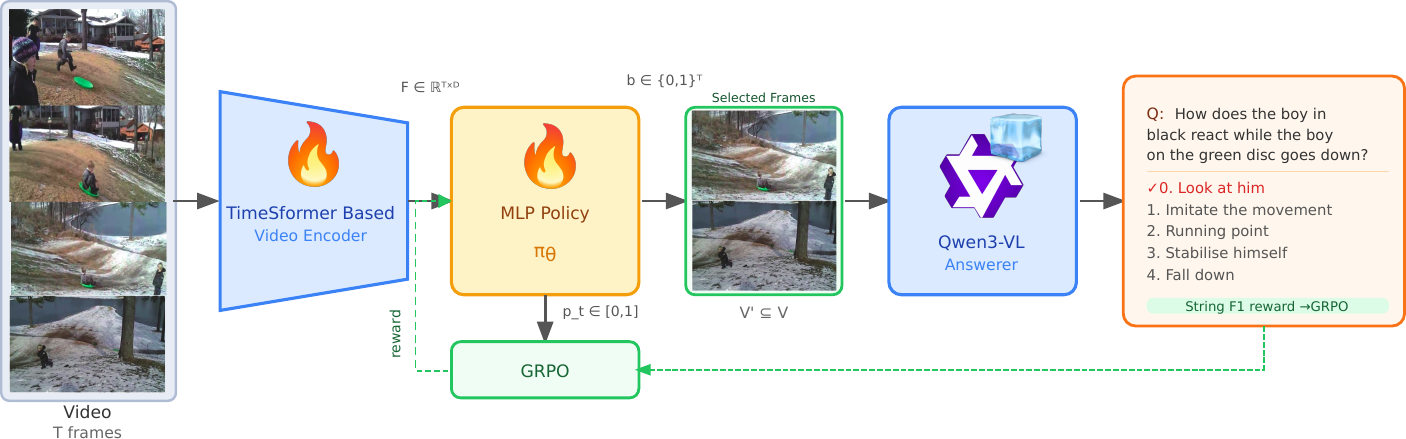}
    \caption{
        \textbf{HORNet pipeline.}
        Given a video $\mathbf{V} = \{v_1, v_2, \ldots, v_T\}$ with $T$ uniformly sampled frames, our TimeSFormer-based video encoder $E$ extracts per-frame features $\mathbf{F} \in \mathbb{R}^{T \times D}$.
        A lightweight trainable \textbf{MLP policy} $\pi_\theta$ scores each frame
        independently, producing keep probabilities $p_t \in [0,1]$ and a binary
        selection mask $\mathbf{b} \in \{0,1\}^T$.
        Only the frames selected by the mask ($\mathbf{V'} \subseteq \mathbf{V}$)
        are passed to the frozen \textbf{Qwen3-VL} answerer.
        For example, to answer \emph{``How does the boy in black react while the
        boy on the green disc goes down?''}, HORNet selects only the frames
        capturing the key interaction moment, discarding irrelevant context;
        correctly predicting \emph{``Look at him''}.
        At training time, \textbf{GRPO} samples $K$ candidate subsets, evaluates
        each via String F1 reward against the ground-truth answer, and updates
        $\pi_\theta$ through group-normalized policy gradients.
        The VLM answerer remain frozen throughout while the encoder and MLP policy are trainable.
    }
    \label{fig:pipeline}
\end{figure*}
}

\newcommand{\encoder}{
\begin{figure}[t]
    \centering
    \includegraphics[width=0.8\linewidth]{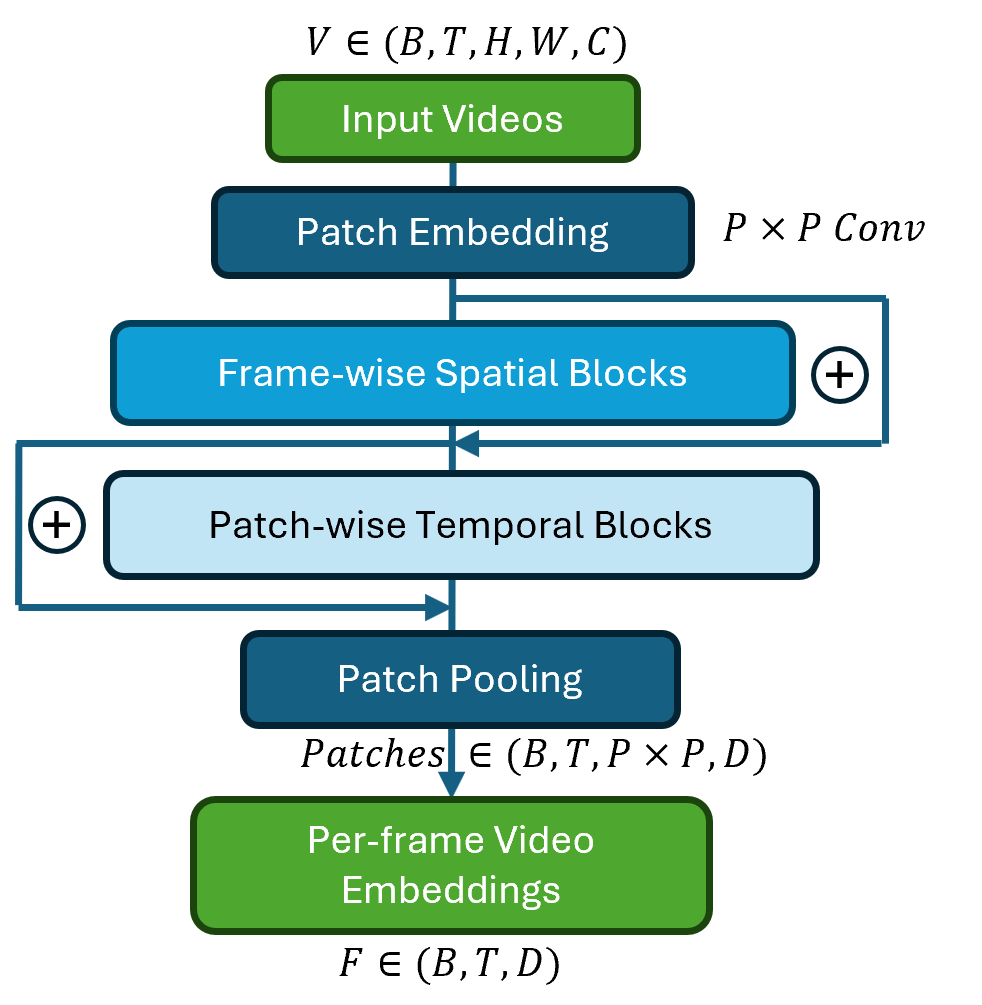}
    \caption{
        \textbf{HORNet encoder $E$.} Input frames are patchified with a $P\times P$ convolution, processed by spatial self-attention within each frame, and then by temporal self-attention across frames at each patch location. The resulting temporally contextualized patch tokens are pooled to yield per-frame video representations used by HORNet for frame selection. $B$ is batch size, $T$ is frame count and $D$ is hidden dimension. We set $P$=16, $T$=32 and $D$=768 in our training.
    }
    \label{fig:encoder}
\end{figure}
}

\newcommand{\quality}{
\begin{figure*}[t]
    \centering
    \includegraphics[width=\linewidth]{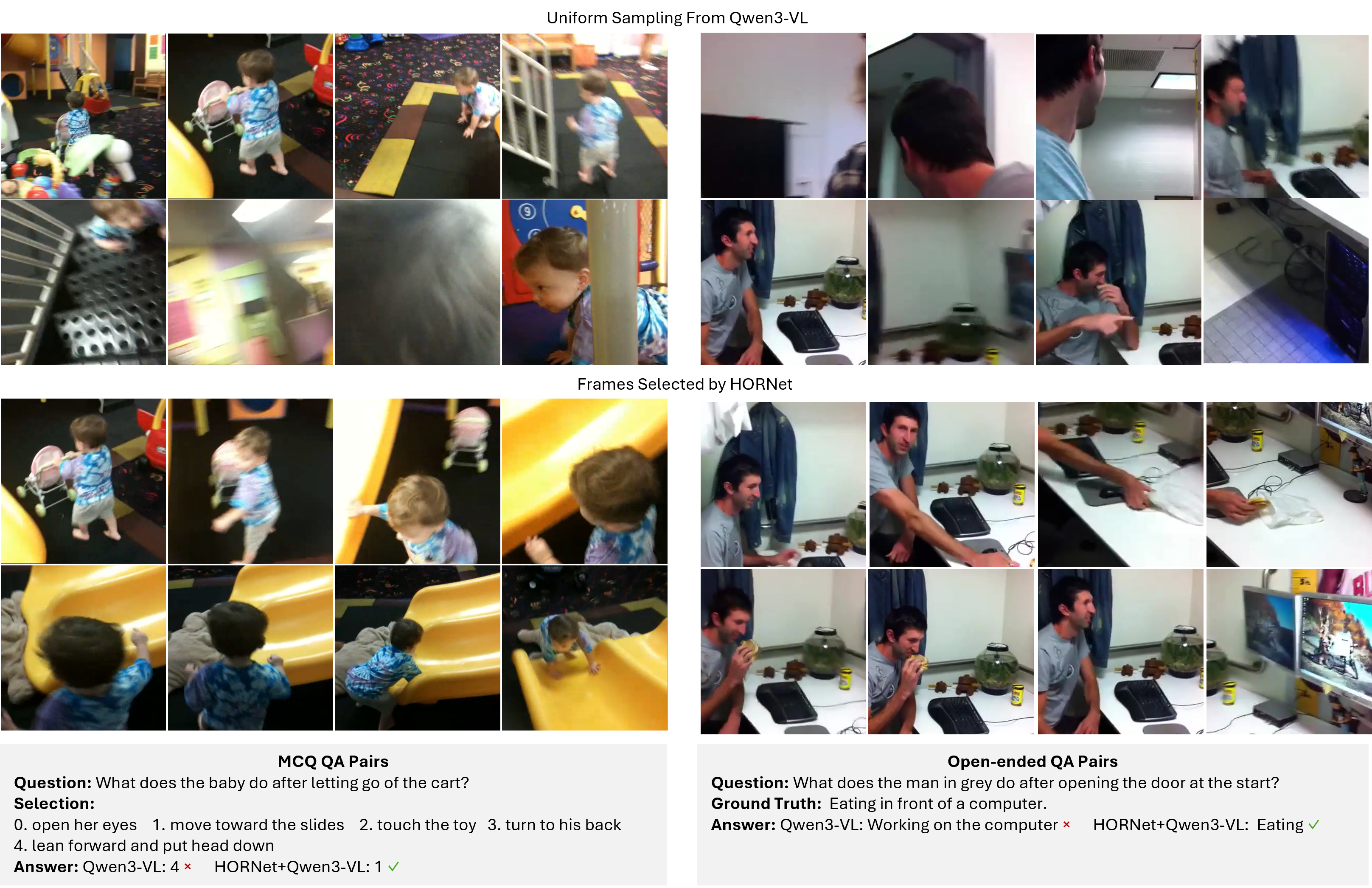}
    \caption{
        Qualitative example of HORNet’s frame-selection behavior on an MCQ and open-ended sample from the NExT-QA dataset \cite{nextqa}. Given fixed 8-frame input, uniform sampling in Qwen-VL captures frames of the child crawling instead of the slide following the discard of the cart (left), and a frame of a person working at a computer while missing the eating frames (right), leading the model to produce an incorrect answer. With a dense initial sampling ($T$=256 frames), HORNet selects the full 8-frame sequence of action-relevant frames while discarding distractors, enabling the model to recover the correct prediction.
    }
    \label{fig:quality}
\end{figure*}
}

%% file: sec/0_abstract.tex
\begin{abstract}
Video question answering (VQA) with vision-language models (VLMs) depends critically on which frames are selected from the input video, yet most systems rely on uniform or heuristic sampling that cannot be optimized for downstream answering quality. We introduce \textbf{HORNet}, a lightweight frame selection policy trained with Group Relative Policy Optimization (GRPO) to learn which frames a frozen VLM needs to answer questions correctly. With fewer than 1M trainable parameters, HORNet reduces input frames by up to 99\% and VLM processing time by up to 93\%, while improving answer quality on short-form benchmarks (+1.7\% F1 on MSVD-QA) and achieving strong performance on temporal reasoning tasks (+7.3 points over uniform sampling on NExT-QA). We formalize this as Select Any Frames (SAF), a task that decouples visual input curation from VLM reasoning, and show that GRPO-trained selection generalizes better out-of-distribution than supervised and PPO alternatives. HORNet's policy further transfers across VLM answerers without retraining, yielding an additional 8.5\% relative gain when paired with a stronger model. Evaluated across six benchmarks spanning 341,877 QA pairs and 114.2 hours of video, our results demonstrate that optimizing \emph{what} a VLM sees is a practical and complementary alternative to optimizing what it generates while improving efficiency. Code is available at \url{https://github.com/ostadabbas/HORNet}.
\end{abstract}

%% file: sec/1_intro.tex
\section{Introduction}
\label{sec:intro}

Existing state-of-the-art VLMs rely on scaling large visual-text data pairs to improve performance~\cite{chen2024expanding,wang2024qwen2vl,liu2023improvedllava,galoaa2026structured}, and while these efforts have yielded measurable gains on VQA benchmarks, the underlying mechanism--a vision encoder tokenizes image patches, a projection layer maps them into the language model's embedding space, and an autoregressive LLM decodes the response--has remained largely unchanged. Videos are first sampled and transformed into visual tokens, which are then aligned with textual inputs through cross-attention mechanisms \cite{galoaa2025lang2motion}. The data-hungry nature of such architecture brings significant downfalls in ``Small Data'' domains, where data collection is costly, inefficient and sometimes facing regulations, limiting the adopting in these situations \cite{ostadabbas2025special,song2026overcoming, bai2023evaluation}. Some approaches attempt to enhance spatial and temporal reasoning by applying LoRA-based~\cite{hu2022lora} fine-tuning on datasets specifically curated for reasoning tasks, thereby improving a model’s ability to understand and interpret video content. Most other methods rely on minor modifications to the attention architecture to adapt the model to specific domains and applications \cite{huang2023posture}. 
\pipelinefig
VLMs for video largely inherit the architecture and biases of image-based models, with temporal reasoning added only superficially. Video-LLaVA~\cite{lin2024videollava} samples eight frames through a shared frozen encoder with no temporal module, LLaVA-OneVision~\cite{li2024llavaonevision} treats video frames as multiple images and shows that an image-only checkpoint already performs competitively on video benchmarks, and Video-ChatGPT~\cite{maaz2024videochatgpt} reduces temporal reasoning to mean-pooling of per-frame CLIP features. The resulting performance gap is stark: InternVL2.5-78B achieves 95.1\% on DocVQA~\cite{mathew2021docvqa} yet only 72.1\% on Video-MME~\cite{chen2024expanding}; a 23-point drop that reflects the absence of genuine temporal reasoning rather than mere task difficulty. In practice, most systems rely on pragmatic frame-sampling strategies, effectively reducing videos to sets of isolated images, leading to unavoidable information loss while increasing signal-noise ratio. The frames with necessary information for VLM to reason might be discarded through this sampling procedures, degrading answers' quality. The broader question of how visual inputs should be structured, represented, and processed within VLMs to preserve key information while removing noise remains insufficiently examined. Although a few studies highlight the role of sampling as a form of filtering~\cite{buch2022revisiting,f2c2025,air2025}, its importance is largely overlooked. 

Given that scaling data has improved image understanding far more than video understanding under this paradigm, we turn to a complementary axis of improvement: optimizing how models reason over their visual inputs through reinforcement-learning-based fine-tuning. Most existing VLMs requires supervised fine-tuning (SFT) to adapt to new domains, which are costly and inefficient with small data. The advent of Group Relative Policy Optimization (GRPO)~\cite{shao2024deepseekmath,deepseekr1} has opened a new avenue for end-to-end optimization of language model behavior via verifiable reward signals. Inspired by its success in guiding the gradients to improve the \emph{outputs} of VLMs~\cite{videor1,r1vl,deepvideor1}, we ask a fundamentally different question: can GRPO be used to optimize \emph{what a VLM sees (inputs)}, rather than what it says (gradients)?
We formalize the problem as  Select Any Frames (SAF): given a video and a question, select the subset of frames from the full temporal sequence that maximizes the downstream VLM's ability to produce the correct answer. SAF treats frame selection as a sequential decision problem amenable to reinforcement learning, with the VLM's QA accuracy providing a direct, task-grounded reward signal. This framing is intentionally simple and general; the SAF policy is modular and can be paired with any downstream VLM without any modifications to the architecture.

To address the issue, we present HORNet: Hindsight Optimization Reasoning, a three-stage SAF pipeline that optimize VLMs' performance by selecting optimal frames from video input (see Fig. \ref{fig:pipeline}). First, a trainable lightweight video encoder extracts rich spatiotemporal features for each frame independently. Second, a lightweight trainable multilayer perceptron (MLP) policy consumes these features and outputs a per-frame keep probability. Third, GRPO trains the video encoder and MLP by sampling multiple candidate frame subsets per video, passing each to a frozen VLM model for answering and computing rewards. Only the MLP and video encoder are trained; the VLM remain frozen throughout. This design makes HORNet exceptionally parameter-efficient-suitable for small-data settings where full fine-tuning of large models is infeasible-while still benefiting from the representational capacity of pretrained video and language foundations. Operating at the frame-selection level rather than the token level allows HORNet to substantially reduce both the memory footprint and the inference latency of downstream VLM processing, with these gains becoming even more pronounced as model size increases. 
% This design offers a meaningful step toward efficient, scalable, and distributed video understanding suitable for edge‑device deployment.

We train HORNet on a diverse multi-dataset corpus spanning MSRVTT-QA~\cite{msrvttqa}, MSVD-QA~\cite{msvdqa}, and NExT-QA~\cite{nextqa}, totaling 17,350 videos, 341,877 QA pairs, and 114.2 hours of content. This breadth covers descriptive, causal, and temporal question types, pushing the policy to discover generalizable selection strategies rather than dataset-specific shortcuts. In short, our contributions are:

\begin{itemize}
    \item We introduce \textbf{SAF (Select Any Frames)}, a task formulation that decouples frame selection from VLM reasoning and enables direct reward-based optimization of visual inputs.
    \item We propose \textbf{HORNet}, a GRPO-trained frame selection policy built on frozen video and language foundations, trainable with minimal parameters.
    \item We demonstrate that GRPO can be redirected from optimizing VLM outputs to optimizing VLM inputs; a conceptual shift that is both more parameter-efficient and more generally applicable.
    \item We provide a large-scale training benchmark combining three VideoQA datasets (341,877 QA pairs, 114.2 hours) for evaluating frame selection methods.
\end{itemize}

\subsection{Small Data Statement}
This work qualifies as small data research on two fronts. First, HORNet is designed for settings where annotated video-question-answer data is scarce or expensive to collect. Rather than fine-tuning a billion-parameter VLM; which typically requires hundreds of thousands of domain-specific examples, HORNet trains fewer than 1M parameters, meaning that the method can be deployed in domains where only a small number of labeled video QA pairs are available, such as medical procedures, surveillance, or industrial inspection, without risking catastrophic forgetting or overfitting of the foundation models.
Second, HORNet's training strategy is explicitly chosen to maximize sample efficiency. GRPO generates multiple candidate frame selections per training example and computes rewards from the frozen VLM's own outputs, effectively amplifying each labeled sample by a factor of $K=8$ without requiring any additional annotation. Our ablation study (Table~\ref{tab:ablation_objective}) confirms this advantage. Furthermore, the trained policy transfers to a different VLM answerer without retraining (Table~\ref{tab:ablation_vlm}), eliminating the need to recollect data when the downstream model changes. Together, these design choices--frozen foundations, reward amplification, and transferable policies--make HORNet particularly suited to the data-scarce regimes that motivate this work.

%% file: sec/2_related.tex
\section{Related Work}
\label{sec:related}
\researchgap
We summarize the positioning of existing methods in Table~\ref{tab:gap} across four desirable properties: whether the method uses learned frame selection, whether it is optimized via downstream reward signals, whether the VLM remains frozen during training, and whether it is parameter-efficient. Existing approaches satisfy at most two of these properties simultaneously. HORNet is the first to satisfy all four.
\paragraph{Frame selection for video understanding.}
The importance of selecting the right frames, rather than sampling uniformly, has been recognized since Buch~\etal~\cite{buch2022revisiting} demonstrated that a single well-chosen frame, identified by a permutation-invariant attention module over frozen CLIP embeddings, suffices for many VideoQA benchmarks. This finding motivated a line of work on learned selection. SeViLA~\cite{sevila} chains a Localizer and Answerer fine-tuned from BLIP-2, using pseudo-labels from the Answerer to self-refine the Localizer. Frame-Voyager~\cite{framevoyager} enumerates frame combinations and trains a supervised selector by ranking subsets according to a Video-LLM's prediction loss. VidF4~\cite{vidf4} proposes differentiable frame scoring that jointly considers question relevance and inter-frame diversity. On the training-free side, F2C~\cite{f2c2025} segments videos into temporally coherent clips using watershed-based scoring and CLIP query relevance, demonstrating that clip-level temporal coherence can outperform isolated frame selection. A.I.R.~\cite{air2025} employs a VLM to iteratively decompose queries and evaluate small frame batches, trading inference cost for selection accuracy. BOLT~\cite{bolt2025} and Q-Frame~\cite{qframe2025} also use CLIP similarity with different sampling strategies to balance query relevance and coverage. These methods demonstrate that the \emph{when} of frame selection matters as much as the \emph{how}. HORNet differs from all of these in that it \emph{learns} the selection policy end-to-end from downstream QA rewards, without heuristics, pseudo-labels, or combinatorial enumeration.

\paragraph{Reinforcement learning for frame selection.}
A concurrent wave of work applies RL specifically to visual input selection. ReFoCUS~\cite{refocus2025} trains an autoregressive frame selector using reward signals from a reference VLM's answer confidence margins. ViaRL~\cite{viarl2025} co-evolves a frame selector and answerer via iterated amplification RL, achieving strong results on temporal needle QA tasks. FrameMind~\cite{framemind2025} introduces Frame-Interleaved Chain-of-Thought with a GRPO variant for multi-turn dynamic resolution frame sampling. VideoBrain~\cite{videobrain2025} trains an agent that decides when to invoke additional frame sampling using GRPO at the agent-invocation level. While these methods share our motivation, they differ in key respects: ReFoCUS uses autoregressive selection; ViaRL modifies both selector and answerer; FrameMind requires multi-turn agentic inference; and VideoBrain operates at the coarse sampling decision level rather than per-frame scoring. HORNet is simpler by design; a single forward pass through a frozen encoder followed by an MLP produces selection probabilities, and GRPO training requires no modifications to either the encoder or the VLM. This simplicity makes it particularly suited to small-data and resource-constrained settings.

\paragraph{GRPO for vision-language models.}
Group Relative Policy Optimization~\cite{shao2024deepseekmath} was introduced to train language models on verifiable rewards without a critic network, later scaled in DeepSeek-R1~\cite{deepseekr1} to incentivize emergent reasoning. Its application to vision-language models has since expanded rapidly: Video-R1~\cite{videor1} applies temporal contrastive GRPO to video MLLMs; R1-VL~\cite{r1vl} extends it to step-wise multimodal reasoning; DeepVideo-R1~\cite{deepvideor1} addresses the vanishing advantage problem specific to video GRPO; Vision-R1~\cite{visionr1} demonstrates data-efficient GRPO training for visual math reasoning; and GRPO-CARE~\cite{grpocare} addresses reasoning consistency degradation. Critically, all of these works apply GRPO to improve what the VLM \emph{generates}; optimizing output distributions. HORNet redirects GRPO toward optimizing what the VLM \emph{receives}; a complementary direction that has not been explored prior to this work.

HORNet sits at the intersection of these three threads. It inherits the GRPO optimization framework from the reasoning literature, the select-then-answer pipeline from the frame selection literature, and the frozen foundation model paradigm from efficient video VLMs. The key novelty is using GRPO's group-relative advantage estimation-critic-free, scalable, and reward-agnostic-to directly maximize downstream QA performance through frame selection, with a parameter footprint small enough for low-data regimes.

%% file: sec/3_method.tex
\section{Method}

\label{sec:method}
In this section, we first formally define the Select Any Frames (SAF) problem. 
We then introduce the HORNet architecture and detail the GRPO-based training procedure.

\subsection{Problem Formulation}
Let $\mathbf{V} = \{v_1, v_2, \ldots, v_T\}$ denote a video represented by $T$ uniformly sampled frames, where $v_t \in \mathbb{R}^{H \times W \times C}$ is the $t$-th RGB frame. 
Let $q$ be a natural language question and $a$ the corresponding ground-truth answer. We denote by $\mathcal{D}$ a dataset of triplets $(\mathbf{V}, q, a)$.
A video encoder $E$ extracts spatiotemporal per-frame representations $\mathbf{F} \in \mathbb{R}^{T \times D}$, from which a lightweight policy selects a subset $\mathbf{V}' \subseteq \mathbf{V}$. A pretrained and frozen VLM $\mathcal{M}$ then produces a predicted answer $\hat{a} = \mathcal{M}(\mathbf{V}', q)$.

The goal of SAF is to learn a parameterized policy $\pi_\theta$ that selects a subset  $\mathbf{V}' = \pi_\theta(\mathbf{V}, q)$ maximizing downstream answering performance. Formally, we seek:
\begin{equation}
\theta^* = \argmax_{\theta} 
\mathbb{E}_{(\mathbf{V}, q, a) \sim \mathcal{D}}
\left[
R\!\left(
\mathcal{M}\!\left(\pi_\theta(\mathbf{V}, q), q\right), a
\right)
\right],
\label{eq:saf}
\end{equation}
where $R(\hat{a}, a)$ is a task-specific reward function measuring the quality 
of the predicted answer $\hat{a}$ relative to the ground-truth $a$ 
(e.g., exact match accuracy). 
The VLM $\mathcal{M}$ remains frozen during training; 
only the policy parameters $\theta$ are optimized.

\paragraph{Policy parameterization.} We represent the policy output as a binary selection mask  $\mathbf{b} = (b_1, \ldots, b_T) \in \{0,1\}^T$, 
where $b_t = 1$ indicates that frame $v_t$ is selected. 
The selected subset is therefore $\mathbf{V}' = \{ v_t \mid b_t = 1 \}.$ The policy defines a distribution over binary masks:
\begin{equation}
    \mathbf{b} \sim \pi_\theta(\mathbf{b} \mid \mathbf{V}, q),
\end{equation}
which factorizes over frames via independent Bernoulli decisions:

\begin{equation}
\pi_\theta(\mathbf{b} \mid \mathbf{V}, q)
= \prod_{t=1}^{T} \text{Bernoulli}(b_t \mid p_t),
\end{equation}
where $p_t \in [0,1]$ is the selection probability for frame $v_t$, 
predicted by the policy network.

This formulation imposes no temporal ordering or contiguity constraints on frame selection, hence the name Select Any Frames (SAF). The policy may therefore learn to select temporally sparse key events,  short critical intervals, or dense motion segments, depending solely on what maximizes the task-driven reward.

\encoder
\subsection{Video Representation} 
We design HORNet to identify the most informative frames in a video while suppressing redundant or noisy content. This process is guided by learned video representations rather than raw pixels, enabling the model to focus on semantically meaningful cues that correlate with downstream performance. To balance representational strength with computational efficiency, HORNet employs a lightweight encoder $E$ derived from the TimeSFormer \cite{bertasius2021space} architecture. The encoder decouples spatial and temporal reasoning into two separate transformer blocks to efficiently model video structure. In spatial blocks, we perform spatial self-attention independently on each of the frames to
capture intra-frame relationships such as object appearance, local motion cues, and spatial layout. After spatial encoding, a second transformer stack performs temporal self-attention to capture motion patterns and temporal dependencies at each patch position. This factorized design (see Figure \ref{fig:encoder}) preserves temporal modeling capacity while avoiding the prohibitive cost of joint attention over all tokens.

\subsection{HORNet Architecture}
HORNet instantiates the SAF policy using three components:  a video encoder, a lightweight trainable policy network,  and a frozen VLM answerer (Fig.~\ref{fig:pipeline}).

% \paragraph{Frozen video encoder.}
% Given a video $\mathbf{V}$ with $T$ frames, we extract frame-level features using a frozen VideoPrism encoder~\cite{videoprism}, pretrained on large-scale video–text corpora. The encoder outputs spatial token maps of shape $T \times P \times P \times D$, where $P=16$ denotes the spatial grid resolution and $D=768$ the feature dimension. 
% To obtain compact per-frame representations, we apply spatial average pooling over the $P \times P$ grid:
% \begin{equation}
% \mathbf{F} = \text{AvgPool}_{2D}\!\left(\text{VideoPrism}(\mathbf{V})\right)
% \in \mathbb{R}^{T \times D},
% \end{equation}
% %
% where $\mathbf{F} = [\mathbf{f}_1,\ldots,\mathbf{f}_T]^\top$ and each $\mathbf{f}_t \in \mathbb{R}^D$ corresponds to frame $v_t$. 
% All encoder parameters remain fixed during training.
\paragraph{Video encoder.}
Given a video $\mathbf{V}$ with $T$ frames, we extract frame-level features using aforementioned enocder and obtain spatial token maps of shape $T \times P \times P \times D$, where $P=16$ denotes the spatial grid resolution and $D=768$ the feature dimension. To obtain compact per-frame representations, we apply spatial average pooling over the $P \times P$ grid:
\begin{equation}
\mathbf{F} = \text{AvgPool}_{2D}\!\left(E(\mathbf{V})\right)
\in \mathbb{R}^{T \times D},
\end{equation}
where $\mathbf{F} = [\mathbf{f}_1,\ldots,\mathbf{f}_T]^\top$ and each $\mathbf{f}_t \in \mathbb{R}^D$ corresponds to frame $v_t$. 

\paragraph{Policy network.}
The SAF policy is parameterized as a frame-wise multilayer perceptron (MLP) that maps each feature vector $\mathbf{f}_t$ to a selection probability $p_t \in (0,1)$.  Specifically, the network applies three linear projections with Gaussian Error Linear Unit (GELU) nonlinearities followed by a sigmoid activation:
\begin{equation}
p_t = \sigma\!\left(
\mathbf{W}_2 \,\phi\!\left(
\mathbf{W}_1 \,\phi\!\left(
\mathbf{W}_0 \mathbf{f}_t
\right)\right)\right),
\end{equation}
where $\phi(\cdot)$ denotes GELU, $\sigma(\cdot)$ the sigmoid function, 
and the weight matrices project $D \rightarrow 512 \rightarrow 256 \rightarrow 1$.  Collectively, these weights define the learnable parameter set $\theta$. 
The resulting probabilities $\mathbf{p} = (p_1,\ldots,p_T)$ define independent Bernoulli decisions over frames, as described in the SAF formulation.  This MLP constitutes the only trainable component of HORNet.

\paragraph{Frozen VLM answerer.}
For a sampled mask $\mathbf{b}$, the selected frames $\mathbf{V}'$ are passed to a frozen Qwen3-VL model~\cite{qwen3vl} together with the question $q$. 
The model produces a predicted answer $\hat{a}$, which is used to compute rewards during training and for evaluation at test time.

\subsection{Training with GRPO}

\paragraph{Candidate generation.}
For each training instance, we generate $K=8$ candidate masks 
$\{\mathbf{b}^{(1)},\ldots,\mathbf{b}^{(K)}\}$. 
Candidates are produced using a deterministic top-$k$ sweep over sorted probabilities $\mathbf{p}$, progressively reducing the number of selected frames, together with one stochastic Bernoulli sample to maintain exploration.

\paragraph{Reward computation.}
Each candidate mask yields a predicted answer 
$\hat{a}^{(i)} = \mathcal{M}(\mathbf{V}'^{(i)}, q)$. 
We define a smooth scalar reward

\begin{equation}
r^{(i)} = 
0.1 \cdot F_1^{\text{token}}(\hat{a}^{(i)}, a)
+ 0.9 \cdot \text{EditSim}(\hat{a}^{(i)}, a),
\end{equation}

where $F_1^{\text{token}}$ denotes token-level F1 after lemmatization and $\text{EditSim}$ is normalized edit similarity in $[0,1]$. 
This formulation reduces brittleness to minor lexical variations.

\paragraph{GRPO objective.}
The log-probability of candidate mask $\mathbf{b}^{(i)}$ under the current policy is

\begin{equation}
\log \pi_\theta(\mathbf{b}^{(i)} \mid \mathbf{F})
=
\sum_{t=1}^{T}
\left[
b_t^{(i)} \log p_t
+
(1 - b_t^{(i)}) \log (1 - p_t)
\right].
\end{equation}

Let $\bar{r}$ and $\sigma_r$ denote the mean and standard deviation of rewards within the group of $K$ candidates. 
The normalized advantage is defined as

\begin{equation}
A^{(i)} = \frac{r^{(i)} - \bar{r}}{\sigma_r + \epsilon},
\end{equation}
where $\epsilon$ is a small constant for numerical stability. 
The GRPO loss is then

\begin{equation}
\mathcal{L}_{\text{GRPO}}
=
-\frac{1}{K}
\sum_{i=1}^{K}
A^{(i)} \,
\log \pi_\theta(\mathbf{b}^{(i)} \mid \mathbf{F}).
\end{equation}

We optimize $\theta$ using Adam with learning rate $10^{-4}$.

%% file: sec/4_results.tex
\section{Results}
\label{sec:results}
In this section, we describe our training data and strategies, and present HORNet’s performance and efficiency gains over the baseline model through both qualitative and quantitative analyses. We conduct ablation studies to examine alternative design choices in VLM architectures, training procedures, and sampling strategies. Overall, we demonstrate substantial efficiency improvements and highlight HORNet’s potential when scaled to larger backbone models.

\subsection{Training Data}
HORNet is trained on a combined corpus spanning three VideoQA benchmarks: \textbf{MSRVTT-QA}~\cite{msrvttqa} (10,000 videos, 158,581 training QA pairs, mean 15.5s duration), \textbf{MSVD-QA}~\cite{msvdqa} (1,161 training videos, 30,933 QA pairs, mean 9.6s), and \textbf{NExT-QA}~\cite{nextqa} (3,870 training videos, 34,132 QA pairs, mean 43.7s). In aggregate, the training set contains \textbf{223,646 QA pairs} across \textbf{15,031 videos} covering 114.2 hours of content, with question types spanning descriptive (what/who), temporal (when/how), and causal (why) reasoning. This breadth ensures the selection policy generalizes across diverse temporal structures rather than overfitting to a single question distribution.

\subsection{Implementation Details}

All experiments are conducted on a single \textbf{NVIDIA A100 40GB} GPU. Videos are decoded and uniformly sampled to $T=32$ frames, each resized to $288 \times 288$ pixels. The \textbf{TimeSFormer-Tiny} encoder produces spatial feature maps of shape $16 \times 16 \times 768$, which are spatially average-pooled to yield per-frame descriptors $\mathbf{F} \in \mathbb{R}^{16 \times 768}$.

The \textbf{MLP policy} $\pi_\theta$ consists of a linear projection ($768 \rightarrow 512$) followed by two hidden layers ($512 \rightarrow 1024 \rightarrow 256$) with GELU activations, and a final linear head ($256 \rightarrow 1$) with sigmoid output. This amounts to fewer than 1M trainable parameters.

At each training step, $K=8$ candidate frame subsets are sampled per video via a top-$k$ sweep with step size $\lfloor k/K \rfloor$. Training proceeds in two stages. In the first stage, we train on MSVD \cite{msvdqa} and MSRVTT \cite{msrvttqa}, which contain short videos (fewer than 100 frames) and one-word answers. Rewards are computed using an F1-Lev objective: a weighted combination of token-level F1 ($w_1$=0.1) and normalized edit similarity ($w_2$=0.9) applied to lemmatized predictions and ground-truth answers. In the second stage, we train on NExT-QA \cite{nextqa}, which features MCQ-style questions and long videos (around 1,000 frames), using a selection-accuracy reward tailored to the multiple-choice setting. The policy is optimized with \textbf{Adam}~\cite{adam} at a learning rate of $10^{-4}$ with batch size 8 on a total of 223,646 training QA pairs. Qwen3-VL-2B stays fully frozen during training, while the video encoder and the frame-selection policy are trained jointly.

\subsection{Open-Ended QA Results}
\mainoe

Table~\ref{tab:mainoe} reports results on three open-ended VideoQA benchmarks. On MSVD-QA, HORNet improves F1-Lev from 0.3483 to 0.3543 (+1.7\%) while reducing Qwen processing time by 64\% and input frames by 66\%. This shows that for short videos ($\sim$10s), many frames are redundant or noisy, and selecting a compact subset actually helps the VLM focus on relevant content.

On MSRVTT-QA and NExT-QA open-ended, HORNet trades a modest drop in F1 for substantial efficiency gains. MSRVTT-QA loses 5.6\% F1 but reduces processing time by 84\% and frames by 92\%. NExT-QA open-ended loses 10.1\% but reduces processing time by 81\% and frames by over 99\%, compressing an average of 1,158 input frames down to 8. These results highlight a practical trade-off: HORNet enables deployment on resource-constrained settings where processing thousands of frames per video is infeasible, with a bounded cost in answer quality.

\subsection{Multiple-Choice QA Results}
\mainmcq

Table~\ref{tab:mainmcq} presents results on three MCQ benchmarks. The pattern mirrors the open-ended setting: HORNet consistently reduces processing time (74--93\%) and frame count ($\geq$99\%) across all datasets. On ActivityNet-QA, accuracy drops only 6.2\% while inference becomes 93\% faster. On NExT-QA MCQ, the gap narrows to 5.3\% with 74\% faster processing. VideoMME shows the largest accuracy gap (16.2\%), which we attribute to its hour-scale videos where 8 frames may be insufficient to cover the question scope.

Across both open-ended and MCQ settings, the results support a consistent finding: HORNet provides a controllable efficiency--accuracy trade-off, achieving order-of-magnitude reductions in computational cost with bounded quality loss. In certain cases, HORNet even improves the model’s predictions by discarding distracting or noisy frames and retaining only the most informative moments, producing a better answer than using VLMs alone, as illustrated in Figure \ref{fig:quality}.

\quality

\subsection{Ablation Studies}

\paragraph{Training objective.}
\ablationobjective
Table~\ref{tab:ablation_objective} compares three training strategies for the frame selection policy, all trained exclusively on MSVD-QA. On the in-distribution MSVD-QA evaluation, all three methods improve over the untrained baseline, with PPO achieving the highest F1 (0.3585) followed by GRPO (0.3543) and SFT (0.3495). However, the MSRVTT-QA column reveals a critical difference: since none of the methods were trained on MSRVTT-QA, this column measures out-of-distribution generalization. Here, all trained policies degrade relative to the untrained baseline (0.3209), but GRPO degrades the least (0.3029), retaining 94\% of baseline performance compared to 92\% for PPO and 90\% for SFT. This suggests that GRPO's group-relative advantage estimation learns more transferable selection strategies, whereas PPO and SFT overfit more aggressively to the training distribution.

\paragraph{Frame selection strategy.}
\ablationselection
Table~\ref{tab:ablation_selection} compares random, uniform, and HORNet selection, all restricted to exactly 4 frames. On MSVD-QA and MSRVTT-QA, all three strategies perform within 0.01 F1 of each other. This is expected: with average durations of 10s and 15s respectively, most frames in these videos carry similar visual content, and any 4-frame sample is likely to capture the relevant information. Notably, the fact that aggressive subsampling (4 out of 32 frames) does not substantially hurt performance reinforces our core premise: many frames are redundant or noisy, and discarding them does no harm.

The picture changes on NExT-QA, where videos average 44 seconds and questions require causal and temporal reasoning. Here, HORNet achieves 71.50\% accuracy, outperforming random (65.88\%) by 5.6 points and uniform (64.24\%) by 7.3 points. When the temporal structure of the video matters, learned selection provides a clear advantage over blind sampling.

\paragraph{VLM answerer.}
\ablationvlm
Table~\ref{tab:ablation_vlm} swaps only the frozen VLM answerer while keeping the same HORNet policy. Replacing Qwen3-VL-2B with the larger Qwen2.5-VL-3B improves F1-Lev from 0.3543 to 0.3846, a 8.5\% relative gain. This confirms that HORNet's frame selection transfers across VLM answerers without retraining, and that pairing the policy with a stronger answerer amplifies the benefit of intelligent frame selection.

%% file: sec/5_conclusion.tex
\section{Conclusion}
\label{sec:conclusion}
We introduced HORNet, a lightweight frame selection policy trained with GRPO that optimizes what a frozen VLM sees rather than what it generates, requiring fewer than 1M trainable parameters. Our experiments show that aggressive frame reduction (to as few as 4 frames) causes no meaningful quality loss on short-form videos, while on longer videos with temporal and causal questions, learned selection outperforms uniform and random baselines by up to 7.3 percentage points. Across all benchmarks, HORNet reduces VLM processing time by 64--93\% and input frames by up to 99\%. Ablation studies further confirm that GRPO generalizes better out-of-distribution than PPO and SFT, and that the learned policy transfers across VLM answerers without retraining, yielding an 8.5\% relative gain when paired with a stronger model.

HORNet's limitation is that the accuracy gap widens on hour-scale videos (e.g., VideoMME), where a fixed budget of 8 frames may be insufficient. Future work could address this through adaptive frame budgets that scale with video duration and hierarchical strategies that first localize relevant temporal segments before selecting frames within them. We also plan to incorporate visual reward signals that directly assess the perceptual quality and informativeness of selected frames, complementing the current text-based QA reward. Additionally, we aim to explore partially unfreezing the VLM answerer so that it can provide gradient-based feedback to the selection policy, enabling a tighter co-optimization loop between frame selection and answer generation.